# COVID-19 forecasting based on an improved interior search algorithm and multi-layer feed forward neural network

Rizk M. Rizk-Allah[1,*], Aboul Ella Hassanien[2,*]

[1]Department of Basic Engineering Science, Faculty of Engineering, Menoufia University, Shebin El-Kom, Egypt
[2]Faculty of Computers and Artificial Intelligence, Cairo University, Egypt
[*]Scientific Research Group in Egypt http://www.egyptscience.net

**Abstract**

COVID-19 is a novel coronavirus that was emerged in December 2019 within Wuhan, China. As the crisis of its serious increasing dynamic outbreak in all parts of the globe, the forecast maps and analysis of confirmed cases (CS) becomes a vital great changeling task. In this study, a new forecasting model is presented to analyze and forecast the CS of COVID-19 for the coming days based on the reported data since 22 Jan 2020. The proposed forecasting model, named ISACL-MFNN, integrates an improved interior search algorithm (ISA) based on chaotic learning (CL) strategy into a multi-layer feed-forward neural network (MFNN). The ISACL incorporates the CL strategy to enhance the performance of ISA and avoid the trapping in the local optima. By this methodology, it is intended to train the neural network by tuning its parameters to optimal values and thus achieving high-accuracy level regarding forecasted results. The ISACL-MFNN model is investigated on the official data of the COVID-19 reported by the World Health Organization (WHO) to analyze the confirmed cases for the upcoming days. The performance regarding the proposed forecasting model is validated and assessed by introducing some indices including the mean absolute error (MAE), root mean square error (RMSE) and mean absolute percentage error (MAPE) and the comparisons with other optimization algorithms are presented. The proposed model is investigated in the most affected countries (i.e., USA, Italy, and Spain). The experimental simulations illustrate that the proposed ISACL-MFNN provides promising performance rather than the other algorithms while forecasting task for the candidate countries.

# 1. Introduction

A novel coronavirus, named COVID-19, emerged in December 2019 from Wuhan in central China. It causes an epidemic of pneumonia in humans and poses a serious threat to global public health. This virus appears through a range of symptoms involving shortness of breath, cough, and fever [1]. Despite the drastic containment measures taken by governments associated with different countries, it is swiftly spread to hit different parts of several countries. Besides China presents the mainland for the outbreak for this epidemic, the USA found itself at the top country with the worst outbreak for this epidemic. The exponential daily increase in the infected people, some governments implements a decree to lockdown entire parts of the country. As the confirmed cases are increased daily and dangerousness of this virus, drastic policies and plans must be explored. In this sense, developing a critical forecasting model to predict the upcoming days is vital for the officials in providing a drastic protection measure.

Recently some efforts have been presented to address the COVID-19. Zhao et al. [2] developed some statistical analysis based on the Poisson concept to expect the real number of COVID-19 cases that had not been reported in the first half of January 2020. They estimated that the unreported cases reach 469 from 1 to 15 January 2020 and those after 17 January 2020 had increased 21-fold. Nishiura et al. [3] presented a statistical estimation model to determine the infection rate regarding the COVID-19 on 565 Japanese citizens (i.e., from 29 to 31 January 2020) which are evacuated from Wuhan, China, on three chartered flights. They estimate the infection rate which is 9.5% and the rate for death which is 0.3% to 0.6%. Tang et al. [4] developed a likelihood-based estimation model to estimate the risk of transmission regarding COVID-19. They concluded the reproduction number can be effectively reduced by the isolation and avoiding the intensive contact tracing. In [5], the transmission risk of COVID-19 through human-to-human is studied on 47 patients. Duccio Fanelli et al. [6] develop a differential equations model to analyze the exponential growth of the COVID-19 on three countries including China, Italy, and France in the time window from 22 / 01 to 15 / 03 / 2020.

Accordingly, the literature involves some models that were developed for forecasting some epidemics. These models include the compartmental model proposed by De. Felice et al. [7] to forecast transmission and spillover risk of the human West Nile (WN) virus. They applied their model on the historical data reported from the mainland of this virus, Long Island, New York, form 2001 to 2014. In [8], forecasting pattern via time series models based on time-delay neural networks, multi-layer perceptron (MLP), auto-regressive, and radial basis function, is proposed to gauge and forecast the hepatitis A virus infection, where these models are investigated on thirteen years of reported data from Turkey country. They affirmed that the MLP outcomes the other models. In [9], a forecasting model using an ensemble adjustment Kalman filter is developed to address the outbreaks of seasonal influenza, where they are employed the seasonal data of New York City from 2003 to 2008. In [10], a dynamic model based on the Bayesian inference concept is presented to forecast the outbreaks of Ebola in some African countries including Liberia, Guinea, and Sierra Leone. Massad et al. [11] developed a mathematical model to forecast and analyze the SARS epidemic while Ong et al. [12] presented a forecasting model for influenza A (H1N1-2009). Moreover, a probability-based model is proposed by Nah et al. [13] to predict the spreading of the MERS.

The feed-forward neural network (FNN) **[14]** presents one of the most commonly used artificial neural networks (ANNs) that has been applied in a wide range of forecasting applications with a high level of accuracy. The FNN possesses some distinguishing features that it makes valuable and attractive for forecasting problems. The first feature is a data-driven self-adaptive technique with few prior assumptions about the model. The second is that it can generalize. The third is that a universal functional approximation that provides a high degree of performance while approximating a large class of functions. Finally, it possesses nonlinear features. Due to these advantages of FNN, it has drawn overwhelming attention in several felids of prediction or forecasting tasks. For example, FNN was presented with two layers for approximating functions [14]. Isa et al. [15] presented the FNN based on multilayer perception (MLP) that is conducted on the data set from the University of California Irvine (UCI) repository. Lin et al. [16] developed a modified FNN based on quantum radial basis function to deal with a dataset from the UCI repository. Malakooti et al. [17] and Hornik et al. [18] proposed the FNN to obtain the optimum solution of the approximation function. In [19,20] the back-propagation (BP) method has been employed as a training technique for FNN. However, these methods perceive some limitations such as falling in a local minimum and slow convergence. To alleviate these limitations, many researchers proposed a combination of meta-heuristic algorithms to improve the performance and the utility of an FNN.

In this context, Zhang et al. [21] developed a combined particle swarm optimization (PSO) based on the BP method. Bohat et al. [22] proposed the gravitational search algorithm (GSA) and PSO for training an FNN while Mirjalili et al. [23] introduced a combination of the PSO and GSA to optimize the FNN, where they introduced an acceleration coefficient in GSA to improve the performance of the overall algorithm. .Si et al. [24] developed a differential evolution (DE)-based method to search for the optimal values of the synaptic weight coefficients of the ANN. Shaw and Kinsner [25] presented the simulated annealing (SA) based on a chaotic strategy to train FNN, to mitigate the possibility of sticking in the local optima, but this algorithm suffered from the slow convergence [26]. Furthermore, Karraboga [27] presented the artificial bee colony (ABC) for training FNN, where it suffers from local exploitation. Irani and Nasimi [28] proposed ant colony optimization (ACO) to optimize the weight vector of the FNN, where it deteriorates the global search. Apart from the previously proposed approaches based on FNN, the literature is very rich with many recent optimization models that have employed for training tasks [29, 30, 31, 32]. Although many related studies seem to be elegant for the forecasting tasks, they may deteriorate the diversity of solutions and may get trapped in local optima. Furthermore, these methods may be problem-dependent. Therefore these limitations can deteriorate the performance of the forecasting output and may achieve the unsatisfactory and imprecise quality of the final outcomes. This motivated us to presents a promising alternative model for forecasting tasks with the aim to achieve more accurate outcomes and avoid the previous limitations by integrating the strengths of a chaotic-based parallelization scheme and interior search method to attain better results.

Interior Search Algorithm (ISA) is a novel meta-heuristic algorithm that is presented based on the beautification of objects and mirrors. It was proposed by Gandomi 2014 [33] for solving global optimization problems. It contains two groups, namely mirror and composition to attain optimum placement of the mirrors and the objects for a more attractive view. The prominent feature of ISA is contained in involving only one control parameter. ISA has been applied to solving many engineering optimization fields [34,35,36,37,38]. However, ISA may face the dilemma of the sucking in the local optimum while implemented for complicated and/or high dimensional optimization problems. Therefore, to alleviate these shortages, ISA needs more improvement strategies to acquire great impacts on its performance.

In this paper, an improved interior search algorithm (ISA) based on chaotic learning (CL), a strategy named ISACL is proposed. The ISACL is started with the historical COVID-19 dataset, and then this dataset is sent to the MFNN model to perform the configuration process based on parameters of the weight and biases. In this context, the ISACL is invoked to improve these parameters as the solutions by starting with ISA to explore the search space and CL strategy to enhance the local exploitation capabilities. By this methodology, it is intended to enhance the quality and alleviate the falling in local optima. The quality of solutions is assessed according to the fitness value. The process of algorithm is continued for updating the solutions (parameters) iteratively until the stop condition is reached, and then the achieved best parameters are invoked for configuring the structure of the MFNN model to perform the forecasting and analyzing the number of confirmed cases of COVID-19. The contribution points for this work can be summarized are as follows:

- A brilliant forecasting model is proposed to deal with the COVID-19 and the analysis for the upcoming days is performed on the basis of the previous cases.
- An improved configuration based-MFNN model is presented ISACL-MFNN algorithm.
- The proposed model is compared with the original MFNN and other meta-heuristic algorithms such as GA, PSO, GWO, SCA, and ISA.
- The performance of the ISACL-MFNN affirms its efficacy in terms of the reported results and can achieve accurate analysis for practical forecasting tasks.

The rest sections of this paper are organized as follows. Section 2 introduces the preliminaries for the MFNN model and the original ISA. Subsequently, the proposed ISACL-MFNN framework is provided for introduced in Section 3. Section 4 shows the experiments and simulation results. Finally, conclusions and remarks are provided in Section 5.

## 2. Preliminaries

This section provides the basic concepts of the multilayer feed-forward neural network and the original interior search algorithm.

### 2.1. Multilayer feed-forward neural network

The artificial neural network (ANN) presents one of widely artificial intelligence methods that are employed for forecasting tasks. Its structure involves the input layer, hidden layers and the output layer. The ANN simulates or look alike the human brain and it contains a number of neurons, where it performs the training and testing scenarios on the input data [39]. Input data in the present work include day and the output data present the daily number of cases. The ANN keeps updating iteratively its network weights that connect the input and output layers to minimize the error among the input data. The ANN has advantages include, it can easily learn along with making decisions and also has the ability to depict a relationship among inputs and the output data without obtaining mathematical formulation. Furthermore, the ANN is easy to implement and the flexibility when it is employed for modeling. However the ANN includes some disadvantages which are, it may generate error while forecasting process, training process may reach unstable results, and involves high dimensions parameters (weights) that are need to be found in optimal manner. Furthermore, low convergence and small sample size issues are two common shortages of ANNs.

To construct the neural network model, the input and output data are determined, then the number of neurons among the number of hidden layers must be carefully chosen because they influence on the training accuracy.

The ANN-based forecasting pattern in the present work uses the MFNN along with two hidden layers. The input layer involves several neurons equal to network inputs (days), where N and M neurons are adopted for the first and second hidden layers, respectively and one neuron has assigned for the output layer **[40,41]**. In this context, the hidden neurons transfer function is the sigmoid function, and transfer function for the output is a linear activation function. The associated output for a certain hidden neuron ( *jth* ) is determined as follows:

$$y_i = \frac{1}{1+e^{-\left(\sum_{i=1}^{n}(v_{ji}x_i - b_j)\right)}}, \quad j = 1, 2, ..., N \tag{1}$$

where $v_{ji}$ represents the weight among the *i*th input neuron and the *j*th first hidden neuron, $x_i$ denotes the *i*th input, $y_j$ defines the first hidden layer output. Here, $b_j$ defines the base of the first hidden layer. Also, the output induced by second hidden layer that is denoted by the symbol $\theta_k$ is calculated as follows.

$$\theta_k = \frac{1}{1+e^{-\left(\sum_{j=1}^{N}(v_{kj}y_j - b_k)\right)}}, \quad k = 1, 2, ..., M \tag{2}$$

where $v_{kj}$ represents the weight among the *j*th first and the *k*th second hidden neurons, $y_j$ denotes the *j*th input, $\theta_k$ defines the second hidden layer output. Here, $b_k$ defines the base of the second hidden layer. The overall output (*OPT*), from the *l*th output layer is computed as follows.

$$OPT_l = \sum_{k=1}^{M}(v_{lk}\theta_k), \quad l = 1, 2, ..., H \tag{3}$$

where $v_{lk}$ denotes the weight for second hidden neuron ($k$th) and the output neuron ($l$th).

The assessment of the algorithm performance while training process is determined by the means of the error that equals the difference between the output of MFNN and the target. Here, the mean square error (MSE) is considered:

$$MSE = \frac{1}{n}\sum_{i=1}^{n}\left[(Y_{actual})_i - (OPT)_i\right]^2 \tag{4}$$

where $n$ defines the number of training patterns, $(OPT)_i$ and $Y_{actual}$ are output obtained by the MFNN and the target (actual) output. Here, the fitness value regarding the training process is computed as follows.

$$Fitness = Min.(MSE) = Min.\left(\frac{1}{n}\sum_{i=1}^{n}\left[(Y_{actual})_i - (OPT)_i\right]^2\right) \tag{5}$$

In this sense, gradient-based back-propagation algorithm is presented to update weights and then minimizes the MSE value among the target output and computed output from the MFNN. Also, the MSE is provided to update the biases through the output layer towards the hidden layers **[40]**. Therefore, the weights and biases can be updated as follows:

$$\begin{aligned} v(t+1) &= v(t) + \Delta v, \quad \Delta v_{ji} = \mu(y_j - x_i) \\ b(t+1) &= b(t) + \Delta b, \quad \Delta b_j = \mu(y_j - x_i) \end{aligned} \tag{6}$$

Where $\mu$ defines the learning rate, $\Delta v_{ji}$ represents the change in the weight which hooks up the inputs of the first hidden neurons, and $\Delta b_j$ denotes the change in bias.

The proposed model considers the days as the input variables for the MFNN model, where during these months; cumulative confirmed infected people by COVID-19 are recorded as the target output of the neural network. As MFNN may suffer from a weak performance while searching process, it can liable to fall into local optima. To solve this dilemma, some meta-heuristic algorithms can be usually used to improve the performance of the MFNN model. In this study, an improved interior search algorithm based-chaotic learning (CL) strategy (ISACL) is introduced to enhance the learning capability of the MFNN network.

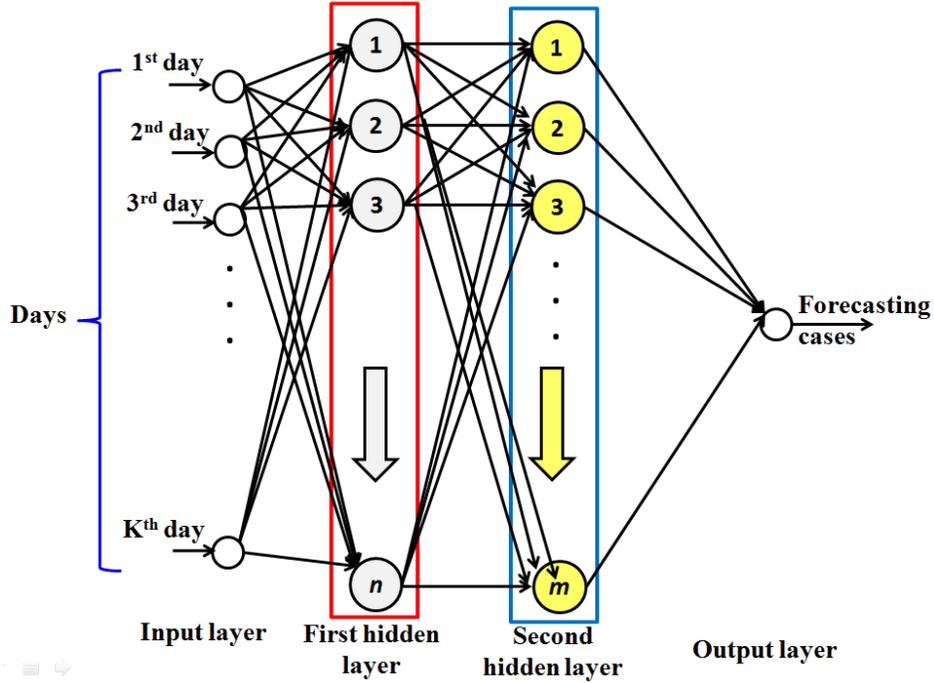

Fig. 1. Structure of the presented MFNN pattern for forecasting task

### 2.2. Interior search algorithm

(1) ISA involves two main stages which simulate the architecture of decoration and interior design. The first stage defines the composition stage in which the composition of elements (solutions in terms of optimization viewpoint) is altered to attain a more beautiful environment, which represents the better fitness in terms of optimization viewpoint. The final one represents a mirror search that aims to explore better views between elements of this stage and the fittest one. More details about ISA are described as follows.

(2) Generate a population of elements randomly between the search bounds, upper ($\Delta_{upper}$) and lower bounds ($\Delta_{lower}$), and record the fitness value for each element.

(3) Obtain the fittest element which is corresponding to minimum objective function in case of minimization. The fittest element is denoted by $\Delta_{gb}^{k}$ for $k$th iteration, where the suffix $gb$ defines the global best.

(4) Divide the population elements into two groups with a random manner, composition group (CG) and mirror group (MG). This can be accomplished by defining the parameter $\lambda$. For the $i$th element, if $r_i \leq \lambda$, then the mirror stage is performed or else it carries out the composition group. Here $r_i$ denotes a random value ranged from 0 to 1.

(5) For composition respect, each element is updated at a random manner within a limited search space and this stage can be formulated as.

$$\Delta_i^k = \Delta_{lower}^k + r_2 \cdot \left( \Delta_{upper}^k - \Delta_{lower}^k \right) \tag{7}$$

where $\Delta_i^k$ defines the $i$th elements of the $k$th iteration and $r_2$ dedicates a random value ranged from 0 to 1.

(6) For the mirror group, the mirror is invoked among each element and the better one (global best). Therefore, the position for $i$th element at $k$th iteration of a mirror is expressed as follows.

$$\Delta_{m,i}^k = r_3 \cdot \Delta_i^{k-1} + (1-r_3) \cdot \Delta_{gb}^k \tag{8}$$

where $r_3$ defines a random value within the interval 0 and 1. The position of the image or virtual position of the element depends on the mirror location, and this can performed as follows.

$$\Delta_i^k = 2\Delta_{m,i}^k - \Delta_i^{k-1} \tag{9}$$

(7) For enhancing the position of the global best, the random walk is implemented as a local search to make slight change for the global best position. This strategy is formulated as follows.

$$\Delta_{gb}^k = \Delta_{gb}^{k-1} + r_n \times \gamma \tag{10}$$

where $r_n$ represents a vector of random numbers with the same size of $\Delta$ that normally distributed and $\gamma$ is a user defined scaling factor that depends on the search space size. Here, $\gamma$ is taken as $0.01\left( \Delta_{upper}^k - \Delta_{lower}^k \right)$.

(8) Obtained the fitness value for each new position and update its location, if it is revival. This can be considered as

$$\Delta_i^k = \begin{cases} \Delta_i^k & f(\Delta_i^k) < f(\Delta_i^{k-1}) \\ \Delta_i^{k-1} & else \end{cases} \tag{11}$$

(9) The procedures are stopped, if assessment criteria are satisfied, repeat from step 2.

The pseudo code for the traditional ISA is portrayed in Fig. 1

| | |
|---|---|
| 1: | **Input:** $T, PS, \Delta_{lower}, \Delta_{upper}, k=1$ |
| 2: | Initialization: $\Delta_i = \Delta_{lower} + random.(\Delta_{upper} - \Delta_{lower}) \forall i \in PS$ |
| 3: | **While** $k \leq T$ |
| 4: | Evaluation: obtain the $\Delta_{gb}^k$ |
| 5: | **for** $i=1$ to $PS$ **do** |
| 6: | **if** $\Delta_{gb}$ |
| 7: | $\Delta_{gb}^k = \Delta_{gb}^{k-1} + r_n \times \gamma$ |
| 8: | **else if** $r_1 \leq \lambda$ |
| 9: | $\Delta_{m,i}^k = r_3 \cdot \Delta_i^{k-1} + (1-r_3) \cdot \Delta_{gb}^k$ |
| 10: | $\Delta_i^k = 2\Delta_{m,i}^k - \Delta_i^{k-1}$ |
| 11: | **Else** |
| 12: | $\Delta_i^k = \Delta_{lower}^k + r_2 \cdot \left(\Delta_{upper}^k - \Delta_{lower}^k\right)$ |
| 13: | Check the boundaries |
| 14: | **end for** |
| 15: | **for** $i=1$ to $PS$ **do** |
| 16: | Evaluate the objective function: $f(\Delta_i^k)$ |
| 17: | Update the solution as follows: |
| 18: | $\Delta_i^k = \begin{cases} \Delta_i^k & f(\Delta_i^k) < f(\Delta_i^{k-1}) \\ \Delta_i^{k-1} & else \end{cases}$ |
| 19: | **end for** |
| 20: | $k = k+1$ |
| 21: | **end while** |
| 22: | **Output**: $\Delta_{gb}$ |

Fig. 2. The framework of the ISA

## 2.3. Motivation of this work

As the series epidemic of the coronavirus, COVID-19, that is outbreak to hit several countries of the globe and causing a considerable turmoil among the peoples, thus the practical intent of the proposed work is to assist the officials with estimating a realistic picture regarding the time and the epidemic peak (i.e., estimating and forecasting the max. no. of infected individuals by the means of forecasting model) and thus can help in developing drastic containment measures for the officials to avoid the epidemic spreading of this virus. In respect of the proposed methodology, an improved interior search algorithm (ISA) is enhanced with chaotic learning (CL) strategy to improve the seeking ability and avoid trapping in the local optima, named ISACL. The ISACL provides an optimization role in achieving an optimal configuration of the MFNN by training process in terms of tuning its parameters. Accordingly, the simulation results have demonstrated the effectiveness and the robustness of the proposed forecasting model while conducting and investigating the foresting tasks.

## 3. Proposed ISACL algorithm

The proposed ISACL algorithm is developed via two improvements which are the composition group based on individuals' experience to emphasize the diversity of the population, and chaotic learning strategy is carried out on the best solution to improve its quality during the optimization process. The detail behind the ISACL is elucidated as follows.

### 3.1. Composition-based experience strategy

In ISA, the element or the individual of the composition group is updated by a random manner that may deteriorate the acceleration of the algorithm and the population diversity. Hence, an experience strategy is introduced to improve the acceleration and the population diversity. In this sense two individuals $\Delta_l^k$ and $\Delta_r^k$ are chosen from the population, randomly. Therefore, the likelihood search direction is illustrated in the updating the current element as follows.

$$\Delta_i^k = \begin{cases} \Delta_i^k + r_2 \cdot (\Delta_r^k - \Delta_l^k) & \text{iff } f(\Delta_r^k) < f(\Delta_l^k) \\ \Delta_i^k + r_2 \cdot (\Delta_l^k - \Delta_r^k) & \text{Otherwise} \end{cases} \tag{12}$$

### 3.2. Chaotic learning based-local searching mechanism

The main merit of the chaos behavior is lies behind the sensitivity to initial conditions, which can potentially perform the iterative search with higher speeds than the conventional stochastic search caused by its ergodicity and mixing properties. To effectively increase the superiority and robustness of the algorithm, a parallelized chaotic learning (CL) strategy is introduced. The CL strategy starts the search with different initial points and thus enhances the convergence rate and overall speed of the proposed algorithm. The steps of CL strategy can be described as follows.

**Step 1. Generation of chaotic values:** In this step, a $N \times D$ matrix $C_k$ of chaotic values is generated according to $N$ maps as follows:

$$C_k = \begin{bmatrix} \alpha_{11}^k & \alpha_{12}^k & \cdots & \alpha_{1D}^k \\ \alpha_{21}^k & \alpha_{22}^k & \cdots & \alpha_{2D}^k \\ \vdots & \vdots & \cdots & \vdots \\ \alpha_{N1}^k & \alpha_{N2}^k & \cdots & \alpha_{ND}^k \end{bmatrix}_{N \times D} \quad (13)$$

where $D$ is the number of dimensions for the decision (control) variable, $\alpha_{jd}^k$ denotes the generated chaotic number within the range of (0, 1) for the $j^{th}$ chaotic map of the $k^{th}$ iteration on the $d^{th}$ dimension. The chaotic numbers in (13) are generated using the functions that are introduced in [42].

**Step 2. Mapping of candidate solution:** For a certain candidate solution, $\boldsymbol{\Delta} = (\Delta_1, \Delta_2, ..., \Delta_D)$, with the $D$ dimensions, the candidate solution can be mapped as follows.

$$X_k^{candidate} = \begin{bmatrix} \boldsymbol{\Delta}_k \\ \boldsymbol{\Delta}_k \\ \vdots \\ \boldsymbol{\Delta}_k \end{bmatrix}_{m \times N} \quad (14)$$

$$X_k^{FIC} = LB + C_k(\Delta_{upper} - \Delta_{lower}) \quad (15)$$

$$Z_k^{chaotic} = \lambda \cdot X_k^{candidate} + (1 - \lambda) X_k^{FIC} \quad (16)$$

where $X_k^{candidate}$ denotes the matrix of the individual $\boldsymbol{\Delta}_t$ repeated $N$ times, $\lambda = k/k_{max}$ identify the weighting parameter and $X_k^{FIC}$ defines the feasible individual that is generated chaotically. In this regard, $X_k^{candidate}$ is considered as the best so for solution.

**Step 3. Updating the best solution:** If $f(Z_k^{chaotic}) < f(\Delta_{gb}^k)$ then put $\Delta_{gb}^k = Z_k^{chaotic}$, otherwise maintains $\Delta_{gb}^k$.

**Step 4. Stopping chaotic search:** If the maximum iteration for the chaotic search phase is satisfied, stop this phase.

### 3.3. Framework of ISACL

Based on the abovementioned improvements, the framework of ISACL is described by the pseudo-code as in Fig.3 and the flowchart in Fig. 4, where the ISACL starts its optimization process by initializing a population of random solutions. Thereafter these solutions are updated by the ISACL and the best solution is refined and updated using the CL phase. Then the superior solution will go to feed the next iteration. The procedures of the framework are continued until any stopping criterion is met.

| | |
|---|---|
| 1: | **Input:** $T, PS, \Delta_{lower}, \Delta_{upper}, k=1$ |
| 2: | Initialization: $\Delta_i = \Delta_{lower} + random.(\Delta_{upper} - \Delta_{lower}) \forall i \in PS$ |
| 3: | **% Phase 1: ISA %** |
| 4: | **While** $k \leq T$ |
| 5: | Evaluation: obtain the $\Delta_{gb}^k$ |
| 6: | **for** $i = 1$ to $PS$ **do** |
| 7: | **if** $\Delta_{gb}$ |
| 8: | $\Delta_{gb}^k = \Delta_{gb}^{k-1} + r_n \times \gamma$ |
| 9: | **else if** $r_1 \leq \lambda$ |
| 10: | $\Delta_{m,i}^k = r_3 . \Delta_i^{k-1} + (1-r_3). \Delta_{gb}^k$ |
| 11: | $\Delta_i^k = 2\Delta_{m,i}^k - \Delta_i^{k-1}$ |
| 12: | **Else** |
| 13: | Select two individuals $\Delta_l^k$ and $\Delta_r^k$ from the population randomly, $\Delta_r^k \neq \Delta_l^k$ |
| 14: | $\Delta_i^k = \begin{cases} \Delta_i^k + r_2.(\Delta_r^k - \Delta_l^k) & \textit{iff } f(\Delta_r^k) < f(\Delta_l^k) \\ \Delta_i^k + r_2.(\Delta_l^k - \Delta_r^k) & \textit{Otherwise} \end{cases}$ |
| 15: | Check the boundaries |
| 16: | **end for** |
| 17: | **for** $i = 1$ to $PS$ **do** |
| 18: | Evaluate the objective function: $f(\Delta_i^k)$ |
| 19: | Update the solution as follows: |
| 20: | $\Delta_i^k = \begin{cases} \Delta_i^k & f(\Delta_i^k) < f(\Delta_i^{k-1}) \\ \Delta_i^{k-1} & else \end{cases}$ |
| 21: | **end for** |
| 22: | $k = k+1$ |
| 23: | **% Phase 2: chaotic learning (CL) strategy %** |
| 24: | Generate $N \times d$ matrix ($C_k$) as in the Eq. (13). |
| 25: | Formulate the $X_k^{candidate}$ and $X_k^{FIC}$ as in the Eqs. (14) and (15) |
| 26: | Map the candidate solution to chaotic space ($Z_k^{chaotic}$) as in the Eq. (16). |
| 27: | If $f(Z_k^{chaotic}) < f(\Delta_{gb}^k)$ then put $\Delta_{gb}^k = Z_k^{chaotic}$ |
| 28: | **end while** |
| 29: | **Output:** $\Delta_{gb}$ |

Fig. 3. The framework of the proposed ISACL

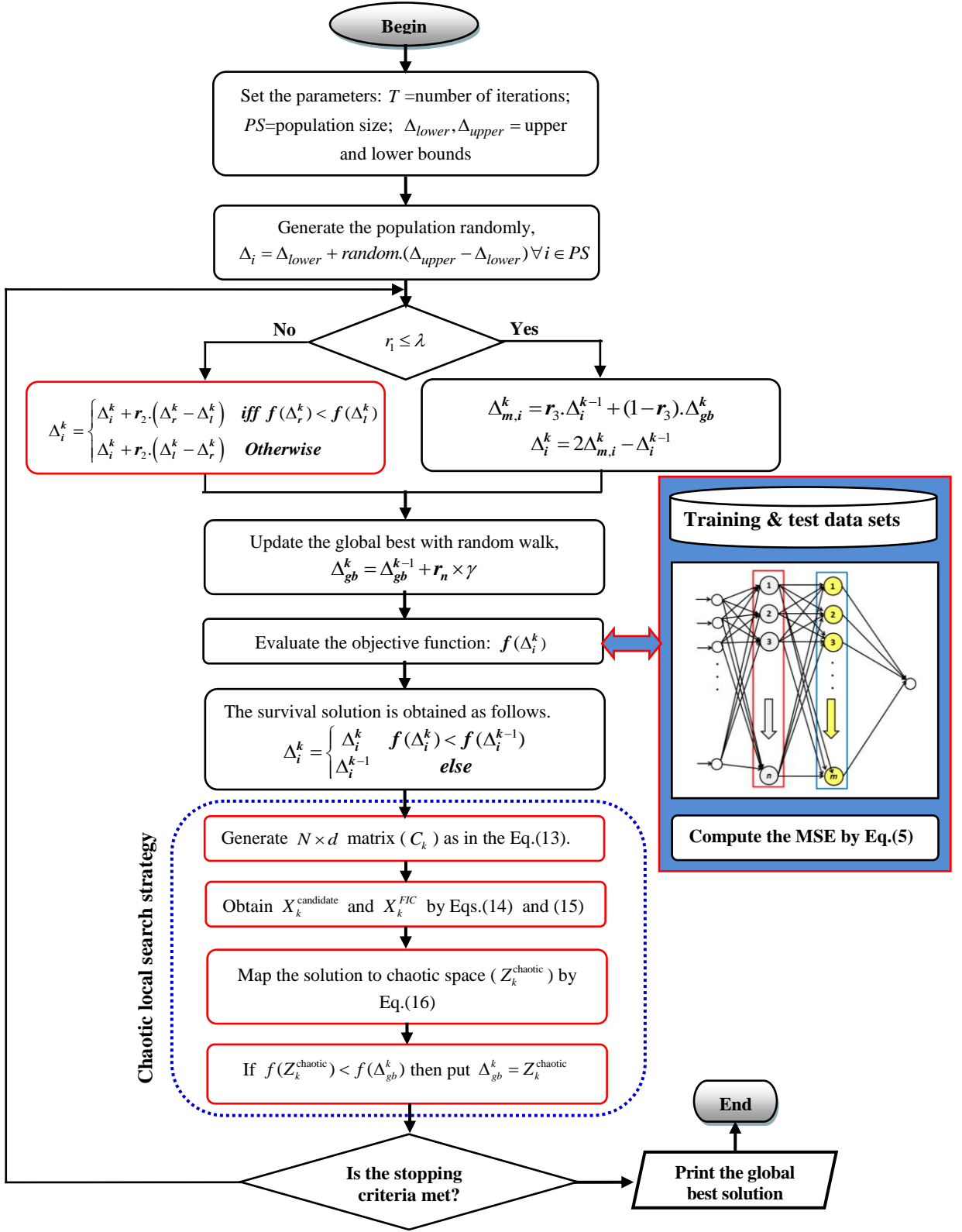

Fig. 4. Training process of the MFNN using the proposed based on the ISACL

## 4. Experiment and Results

In this section, the description regarding the COVID-19 dataset, the parameter settings for the presented algorithms, the performance measures, the results and modeling analysis, and discussions are presented.

### 4.1. Description of the COVID-19 dataset

This subsection provides the COVID-19 dataset that is presented in this study. The COVID-19 dataset is gathered from the WHO website [43] that involved the confirmed people. In this sense, the MFNN is setup with the input data that are represented by the days and output or target that is represented by reported cases, where about 75% the reported cases are employed for training the model while the rest ones are utilized for validating the model and then the model can generalize for any upcoming input cases. In this study, the data of three countries with the larger infected populations are selected, including the USA, Italy, and Spain and referring to the period 22/1/2020 to 3/4/2020.

### 4.2. Parameter Settings

To assess the proposed ISACL-MFNN while forecasting the COVID-19, it is compared with different algorithms including GA [29], PSO [44], GWO [44], SCA [44] and the standard ISA. Because of the high degree of haphazardness associated with the meta-heuristic algorithms and to ensure fairness, each algorithm is carried out with different independent runs and the best result is reported along with performance measures. To attain fair comparisons, the operation parameters such as a maximum number of iterations, populations' size are set to 500 and 10 which are common parameters iterations where the other related parameters associated with each algorithm are provided as reported in its corresponding literature where the overall parameters are tabulated in Table 1. Besides, all algorithms are coded with Matlab 2014b, Windows 7 (64bit) - CPU Core i5 and 4GB RAM.

Table 1 the parameters structures of the implemented algorithms

| Algorithm | Parameters |
|---|---|
| PSO | PS =10, inertia weight: $w_{min} = 0.1$, $w_{max} = 0.4$, acceleration coefficients: $c_1=c_2=2$ |
| GA | PS =10, crossover probability: $cr = 0.25$, mutation probability: $pm = 0.2$ |
| GWO | PS =10, $a = 2-0$, $r_1 = random, r_2 = random, A = 2.a.r_2 - a, C = 2.r_1$ |
| SCA | PS =10, $c_2 \in [0, 2\pi], c_3 \in [0, 2]$, $c_1 = 1-0$ (Linear decreasing) |
| ISA & ISACL | PS =10, tuning parameter: $\alpha = 0.2$ |

### 4.3. Indices for Performance assessment

To further assess the accuracy and quality of the presented algorithms, some performance indices are employed as follows:

4.3.1. Root Mean Square Error (RMSE):

$$RMSE = \sqrt{\frac{1}{N}\sum_{i=1}^{N}\left[(Y_{actual})_i - (Y_{model})_i\right]^2} \tag{17}$$

where $N$ represents the sample size of the data.

4.3.2. Mean Absolute Error (MAE):

$$MAE = \frac{1}{N}\sum_{i=1}^{N}\left|(Y_{actual})_i - (Y_{model})_i\right| \tag{18}$$

4.3.3. Mean Absolute Percentage Error (MAPE):

$$MAPE \frac{1}{N}\sum_{i=1}^{N}\left|\frac{(Y_{actual})_i - (Y_{model})_i}{(Y_{model})_i}\right| \tag{19}$$

4.3.4. Root Mean Squared Relative Error (RMSRE):

$$RMSRE = \sqrt{\frac{1}{N}\sum_{i=1}^{N}\left[\frac{(Y_{actual})_i - (Y_{model})_i}{(Y_{model})_i}\right]^2} \tag{20}$$

4.3.5. Coefficient of Determination ($R^2$):

$$R^2 = 1 - \frac{\sum_{i=1}^{N}\left[(Y_{actual})_i - (Y_{model})_i\right]^2}{\sum_{i=1}^{N}\left[(Y_{actual})_i - \bar{Y}_{model}\right]^2} \tag{21}$$

where $\bar{Y}_{model}$ denotes the mean of $(Y_{actual})_i$ for all $i$.

The smaller values for these metrics (i.e., RMSE, MAE, MAPE, and RMSRE), the higher the accuracy of forecasting model, while the higher value of $R^2$ denotes a high level of correlation. Thus the closer the value of $R^2$ is to 1, the superior result for the candidate method.

Table 2
The results of error analysis regarding the training set (22/1/2020-30/3/2020)

| Countries | Method | RMSE | MAE | MAPE | RMSRE | $R^2$ |
|---|---|---|---|---|---|---|
| **USA** | MFNN | 151887.9 | 135994.9 | 2.279405 | 5.377831 | 0.570935 |
| | GA-MFNN | 20935.81 | 13534.54 | 0.90226 | 0.930784 | 0.787904 |
| | PSO-MFNN | 6269.315 | 4959.217 | 2.136618 | 10.78139 | 0.982571 |
| | GWO-MFNN | 2572.364 | 1772.057 | 0.778763 | 0.861987 | 0.997165 |
| | SCA-MFNN | 40486.76 | 28271.16 | 1.071431 | 1.18745 | 0.264702 |
| | ISA-MFNN | 13132.18 | 10816.57 | 0.88723 | 0.943304 | 0.924713 |
| | ISACL-MFNN | 1217.795 | 986.9662 | 0.960242 | 1.450868 | 0.999345 |
| **Italy** | MFNN | 100629.8 | 93343.47 | 2.887692 | 7.064431 | 0.687694 |
| | GA-MFNN | 5130.22 | 4357.698 | 1.767679 | 8.787012 | 0.983191 |
| | PSO-MFNN | 2023.738 | 1775.697 | 0.862244 | 1.634832 | 0.997424 |
| | GWO-MFNN | 1023.525 | 696.0177 | 0.590222 | 0.806383 | 0.999347 |
| | SCA-MFNN | 8488.579 | 7588.972 | 0.987913 | 2.80546 | 0.973179 |
| | ISA-MFNN | 2757.326 | 1763.483 | 0.794823 | 1.468295 | 0.99566 |
| | ISACL-MFNN | 567.1477 | 289.2544 | 1.095606 | 4.042 | 0.999796 |
| **Spain** | MFNN | 83942.93 | 76222.72 | 2.548526 | 6.236406 | 0.620392 |
| | GA-MFNN | 8345.796 | 6066.738 | 0.821872 | 0.869479 | 0.917202 |
| | PSO-MFNN | 2136.254 | 1421.257 | 1.31129 | 4.27538 | 0.994646 |
| | GWO-MFNN | 3988.714 | 3215.954 | 0.790442 | 0.890076 | 0.981012 |
| | SCA-MFNN | 23524.13 | 16299.57 | 1.132657 | 1.313739 | 0.28774 |
| | ISA-MFNN | 5748.391 | 4498.146 | 0.82158 | 0.923754 | 0.962392 |
| | ISACL-MFNN | 728.2184 | 323.3407 | 0.52082 | 0.666032 | 0.999375 |

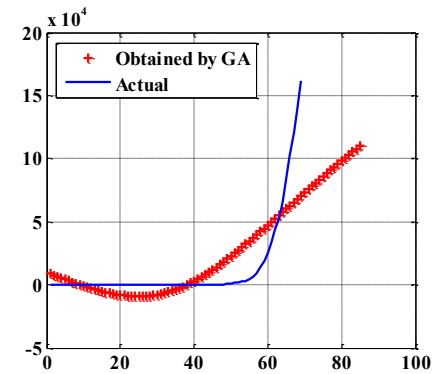
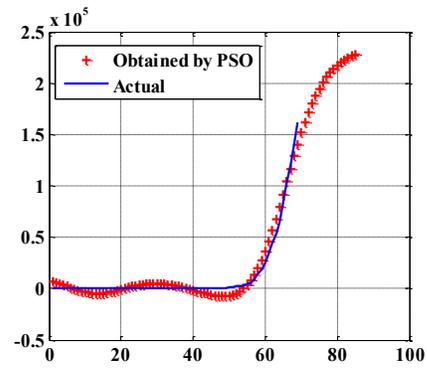
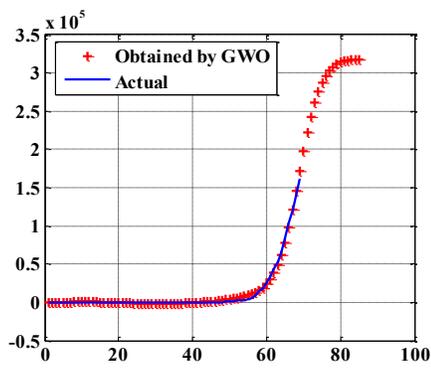
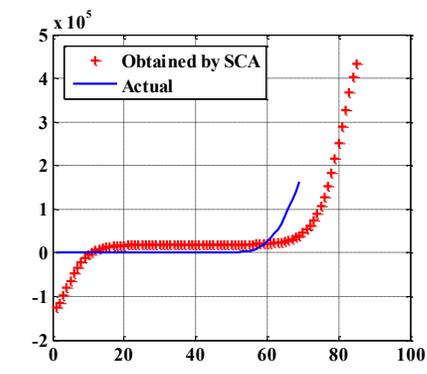

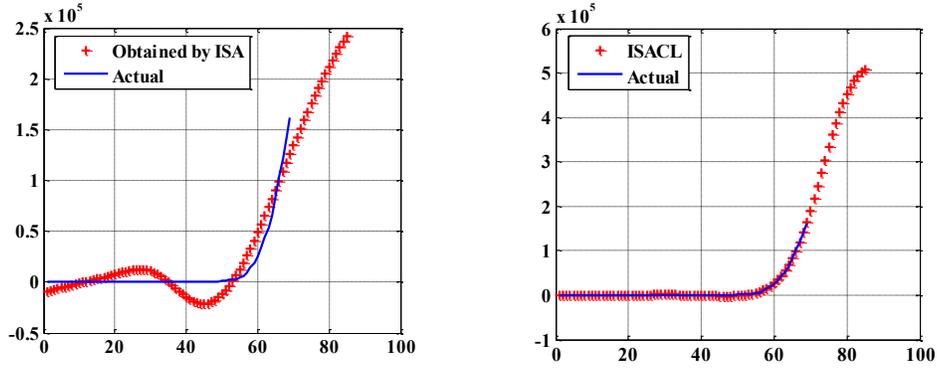

Actual: Days from January 22, 2020 to April 3, 2020 (1:73 days)

Forecasted: April 4, 2020 to April 15, 2020 (74:85 days)

**Fig.5.** The actual data (target) with the obtained (forecasted data) by the presented methods for USA

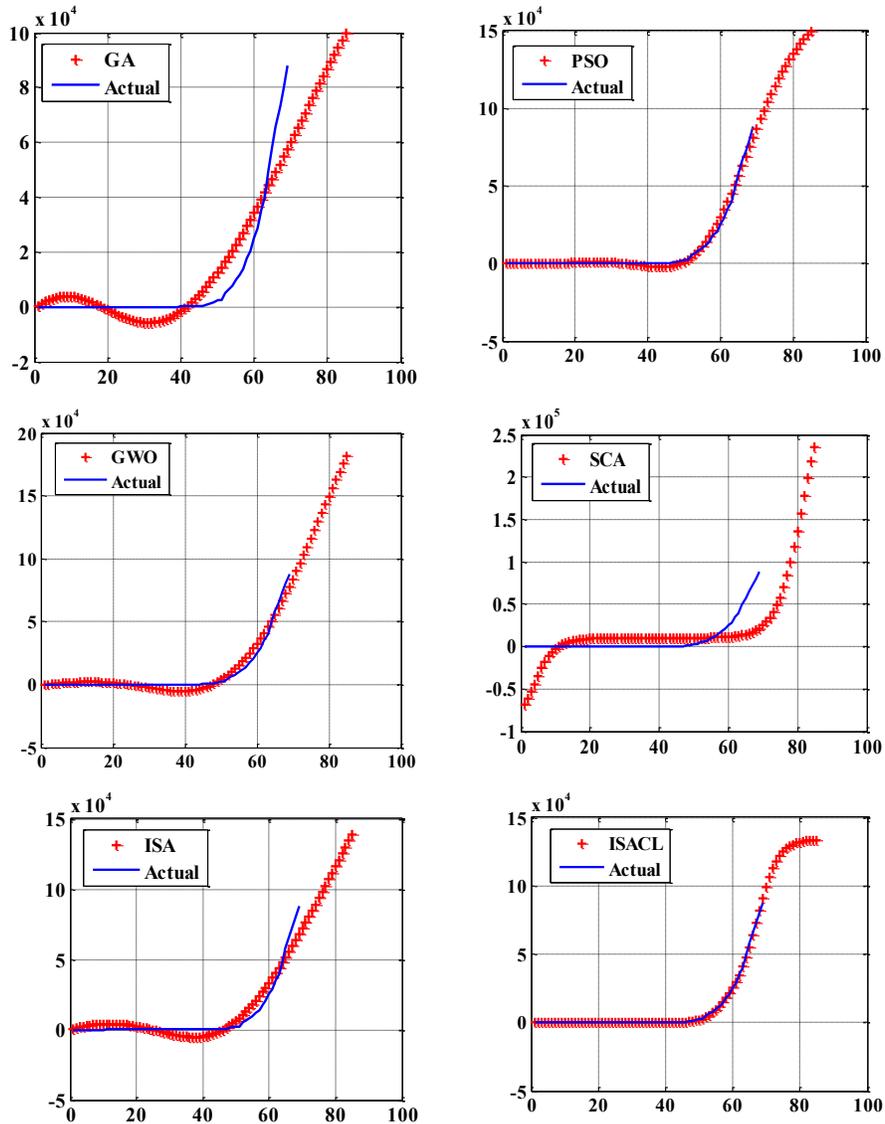

Fig.6. The actual data (target) with the obtained (forecasted data) by the presented methods for Spain

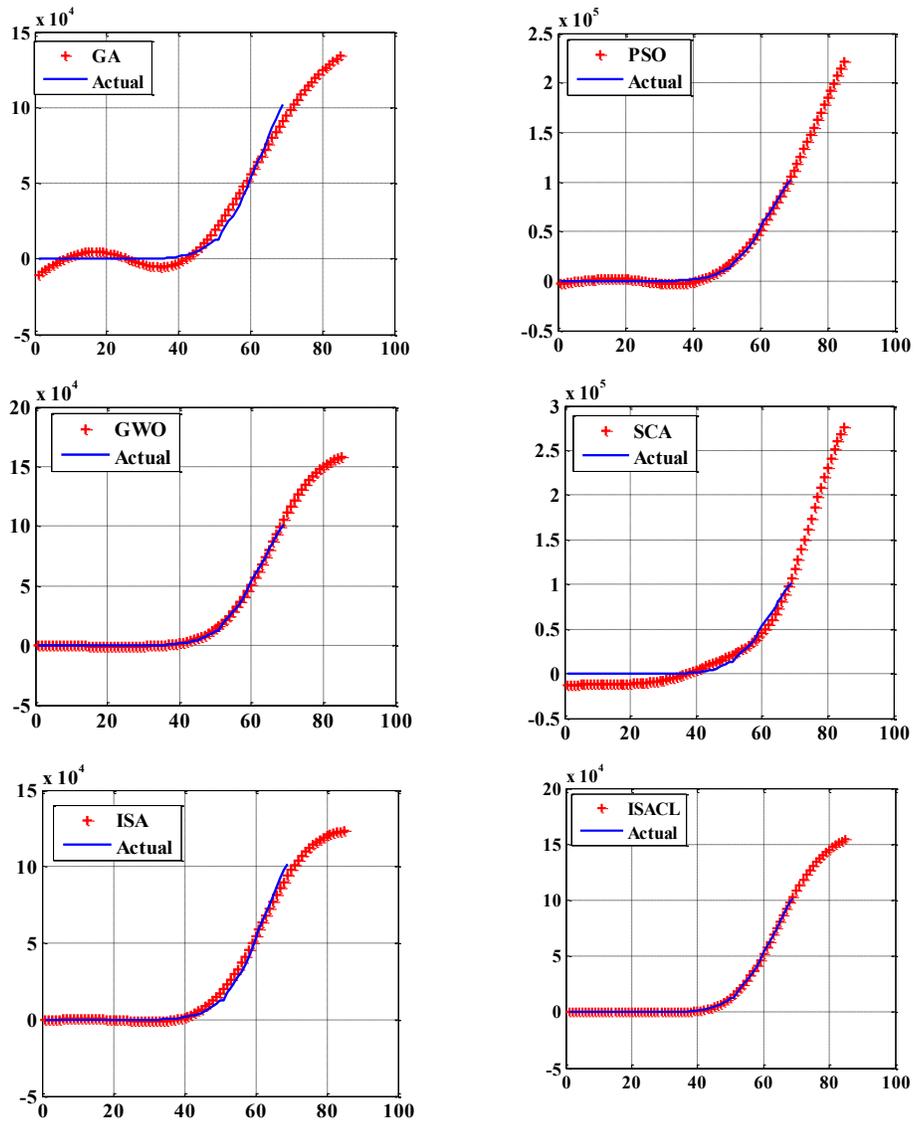

Fig.7. The actual data (target) with the obtained (forecasted data) by the presented methods for Italy

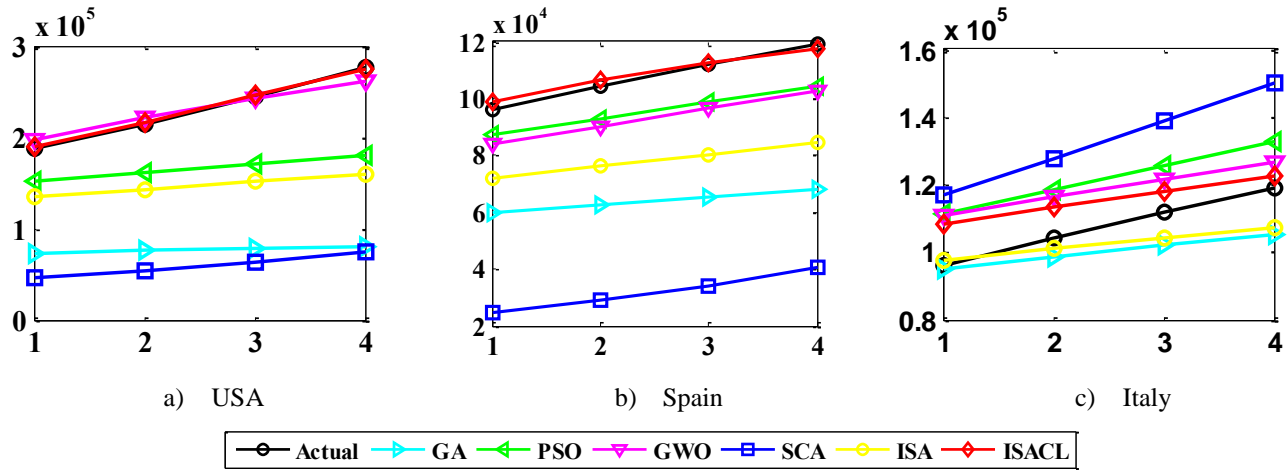

a) USA  b) Spain  c) Italy

Fig. 8. Forecasting results (31/3/2020-3/4/2020) of the confirmed cases for the studied different countries (test set)

Table 3 Forecasting data for coming days of the confirmed cases in the studied countries: 4 / 4 / 2020 − 15 / 4 / 2020.

| Data | Countries | | |
|---|---|---|---|
| | USA | Italy | Spain |
| 4/4/2020 | 304000.7 | 126775.3 | 121728.6 |
| 5/4/2020 | 332772.8 | 130618.7 | 124844.1 |
| 6/4/2020 | 360440.2 | 134156.3 | 127212.2 |
| 7/4/2020 | 386487 | 137392.7 | 128986.4 |
| 8/4/2020 | 410492.9 | 140336.8 | 130301.5 |
| 9/4/2020 | 432150.7 | 143001 | 131268.8 |
| 10/4/2020 | 451271.2 | 145400.6 | 131977.2 |
| 11/4/2020 | 467775.5 | 147552.4 | 132495.3 |
| 12/4/2020 | 481680.6 | 149474.6 | 132875.8 |
| 13/4/2021 | 493078.9 | 151185.7 | 133158.2 |
| 14/4/2021 | 502117.5 | 152704.3 | 133372 |
| 15/4/2021 | 508978.9 | 154048.6 | 133539.6 |

### 4.4. Results and discussions

To forecast the COVID-19 of the confirmed cases of three countries (i.e., USA, Spain, Italy) that are most influenced at 3/4/2020, six optimization algorithms are implemented, where the ISACL presents the proposed one. In this sense, the results of the ISACL are compared with other competitors. The assessments regarding these algorithms are performed by some indices that are reported in Table 2. Based on the reported results, it can be observed that the proposed ISACL outperforms the other models, where it can provide that the lower values for the overall indices including RMSRE, RMSE, MAPE, and MAE and can perceive a higher value for $R^2$ that indicate an accurate correlation between the target and the forecasted results which has nearly 1. On the other hand, the forecasted results obtained by all algorithms for the three countries are depicted in Figs 5-7. These figures depict the training of the presented algorithms via the historical data of the COVID-19 and the forecasting values for twelve days. Also the forecasted cases for the upcoming days starting from 4/ 4 / 2020 to 15 / 4 / 2020 are presented in Table 3. For further validation, the forecasting results of the confirmed cases of the test set , 31/3/2020-3/4/2020, are depicted in Fig. 8, where training set is considered from 22/1/2020 to 30/3/2020.  From Fig. 8, it can be noted that the proposed ISACL is very close to the actual historical data than the other algorithms.

Finally, based on the obtained results, it can be observed that the proposed ISACL-MFNN provides accurate results and has high ability to forecast the COVID-19 dataset regarding the studied cases. In this sense, the limitations of traditional MFNN are avoided due to the integration with the ISACL methodology through the PL strategy to enhance the local exploitation capabilities and obtain high quality of solutions.

## 5. Conclusions

This paper presented an improved interior search algorithm based on chaotic learning (CL) strategy, named ISACL, which is implemented to improve the performance of the MFNN by finding its optimal structure regarding the weights and biases. The developed ISACL-MFNN model is presented as a forecasting model to deal with the novel coronavirus, named COVID-19 that was explored in December 2019 within Wuhan, China. The proposed ISACL-MFNN model is investigated to predict the upcoming days regarding the confirmed, deaths and recovered cases. The performance of the proposed ISACL-MFNN is assessed through the RMSE, MAE, MAPE, RMSRE, and R2 metrics.  The obtained results affirmed the effectiveness and efficacy of the proposed model for predicting task.  Accordingly, the proposed ISACL-MFNN can present a promising alternative forecasting model to deal with practical forecasting applications. To end with, we hope the presented model provides a quantitative picture to help the researchers in a specific country to forecast and analyze the spreading of this epidemic with time.